# Improving Pairwise Ranking for Multi-label Image Classification

Yuncheng Li
University of Rochester
yli@cs.rochester.edu

Yale Song
Yahoo Research
yalesong@yahoo-inc.com

Jiebo Luo
University of Rochester
jluo@cs.rochester.edu

## Abstract

*Learning to rank has recently emerged as an attractive technique to train deep convolutional neural networks for various computer vision tasks. Pairwise ranking, in particular, has been successful in multi-label image classification, achieving state-of-the-art results on various benchmarks. However, most existing approaches use the hinge loss to train their models, which is non-smooth and thus is difficult to optimize especially with deep networks. Furthermore, they employ simple heuristics, such as top-k or thresholding, to determine which labels to include in the output from a ranked list of labels, which limits their use in the real-world setting. In this work, we propose two techniques to improve pairwise ranking based multi-label image classification: (1) we propose a novel loss function for pairwise ranking, which is smooth everywhere and thus is easier to optimize; and (2) we incorporate a label decision module into the model, estimating the optimal confidence thresholds for each visual concept. We provide theoretical analyses of our loss function in the Bayes consistency and risk minimization framework, and show its benefit over existing pairwise ranking formulations. We demonstrate the effectiveness of our approach on three large-scale datasets, VOC2007, NUS-WIDE and MS-COCO, achieving the best reported results in the literature.*

## 1. Introduction

Multi-label image classification is arguably one of the most important problems in computer vision, where the goal is to identify all existing visual concepts in a given image [3]. It has numerous real-world applications including text-based image retrieval [6], ads re-targeting [14], cross-domain image recommendation [35], to name a few. Due to its importance, the problem has been studied extensively, not only in the context of image classification, but from multiple disciplines and in a variety of contexts.

One popular approach is called problem transformation [21], where the multi-label problem is transformed into multiple binary label problems. Several recent approaches exploit some distinctive properties of the problem, such as label dependency [1, 25], label sparsity [10, 12, 27], and label noise [33, 39]. Motivated by the success of deep convolutional neural networks (CNNs) [13, 23], other recent approaches combine representation learning and multi-label learning into an end-to-end trainable system [29].

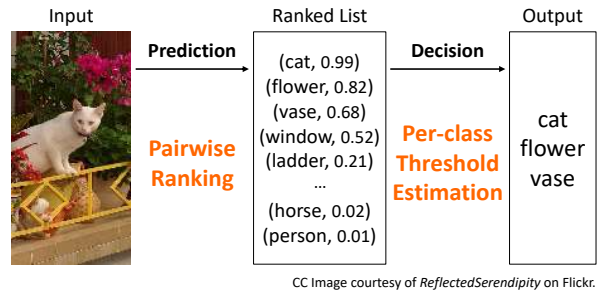

Figure 1: Ranking-based multi-label classification employs a two-step process: label prediction that produces a ranked list of label confidence scores, and label decision that determines which labels to include in the output. We propose a new pairwise ranking loss function and a per-class threshold estimation method in a unified framework, improving existing ranking-based approaches in a principled manner.

Recently, Weston *et al*. [32] proposed an idea to apply pairwise ranking to the image classification problem. Their main idea is that, while what we care the most about is correctly identifying positive labels, it is equally important for the classifier to make "sensible" mistakes. Specifically, even when a classifier fails to identify positive labels, it should at least assign higher ranks to the positive labels than to most of the negative labels. Extending this idea, Gong *et al*. [9] applied the pairwise ranking approach to train a CNN and reported the state-of-the-art result on the NUS-WIDE multi-label image annotation task [7].

While the pairwise ranking approach in Weston *et al*. [32] and Gong *et al*. [9] provides flexibility to train a variety of learning machines, with good empirical performance on real-world problems, we argue that it has two important drawbacks when applied to multi-label classification. First, as we show in Section 3, the hinge loss function used in Weston *et al*. [32] and Gong *et al*. [9] is non-smooth and thus is difficult to optimize. Second, the ranking objective does not fully optimize the multi-label objective.

To see the second drawback at the conceptual level, there are two ways to express the objective of multi-label classification. One is to use the exact match,

$$\min \sum_i \mathbb{I}[\hat{Y}_i == Y_i], \quad (1)$$

where $\hat{Y}_i$ and $Y_i$ are the predicted and the ground-truth labels for the $i$-th example in a dataset, and $\mathbb{I}[\cdot]$ is an indicator function. This considers the prediction to be correct only if it is the same as the ground-truth labels. Another, more relaxed version is to use the Hamming distance,

$$\min \sum_i |\hat{Y}_i \cup Y_i - \hat{Y}_i \cap Y_i|, \quad (2)$$

which minimizes the total number of individual labels predicted incorrectly. Notice how both objectives are different from the ranking objective,

$$\min \sum_i \sum_y \mathbb{I}[rank(y_{pos}) < rank(y_{neg})]. \quad (3)$$

where $y_{pos}$ and $y_{neg}$ are elements in $Y$ for the positive and the negative labels, respectively. It minimizes the number of cases where a positive label has a lower rank than a negative label. Although this is exactly what makes ranking methods achieve its goal, when applied to multi-label classification it lacks the "label decision" part that determines what labels to appear in the prediction result, from the ranked list of label confidence scores (see Figure 1). Previous works used heuristics, such as choose top-$k$ and thresholding, but they ignore image content when making the decision, which is problematic because the number of visual concepts should depend on image content. This discrepancy in the objectives makes ranking approaches somewhat suboptimal for multi-label classification.

In this work, we solve the two aforementioned problems in a principled manner. We propose a novel loss function for pairwise ranking based on a log-sum-exp function. Our loss function is a smooth approximation to the conventional hinge loss, and is smooth everywhere and is easier to optimize. This alone does not solve the absence of label decision in ranking formulation, but allows us to achieve a better ranking objective. Next, we incorporate a label decision module into the classifier, estimating the optimal confidence thresholds for each visual concept appearing in an image.

We provide theoretical analyses of our loss function from the point of view of the Bayes consistency and risk minimization, and show its benefit over existing pairwise ranking formulations. We also demonstrate our method on three large-scale multi-label datasets, VOC2007 [8], NUS-WIDE [7] and MS-COCO [15], and report the best published results in the literature in terms of precision and F1 score. Notably, our method achieves the best performance on the most strict metric, the exact match score in Eqn (1), on all three datasets.

## 2. Related Work

Multi-label classification is a long-standing problem that has been tackled from multiple angles. One common approach is problem transformation. For example, Kim *et al*. [20] treat each label independently and use a binary logistic loss to predict each label, while Read *et al*. [21] transform the multi-label problem into multiple single-label problems using classifier chains and power sets. Zhang *et al*. [38] proposed ML-kNN, using k-nearest neighbor to predict labels for unseen data from training data.

Recent approaches exploit various properties of the multi-label problem, *e.g.*, label dependency [1, 17, 38, 25], label sparsity [10, 12, 27], and label noise [33, 39, 4]. Among these, perhaps the most common approach is to leverage the relationship between labels, *e.g.*, co-occurrence statistics on certain groups of labels. Alessandro *et al*. [1] extended single-label naive Bayes classifier to the multi-label case, proposing an ensemble of Bayesian networks in tree-structured label space. Liu *et al*. [17] proposed to use large-margin metric learning by incorporating label dependency, extending the ML-kNN approach [38]. In the perspective of subspace learning, Shu *et al*. [25] proposed a scalable dimensionality reduction method for multi-label data based on dependency maximization.

Other approaches exploit label sparsity, based on an observation that only a few labels out of, say several thousands, are present in any given instance. Hsu *et al*. [10] proposed a compressed sensing based technique to solve multi-label classification, while Kapoor *et al*. [12] proposed a Bayesian version of compressed sensing. Song *et al*. [27] extended this by incorporating label dependency structures.

Another notable direction is to deal with label noise. For example, Wu *et al*. [33] proposed a mixed graph approach to enforce label consistency and hierarchy structure, effectively handling missing labels; Zhao *et al*. [39] proposed a semi-supervised learning framework based on low rank label matrix recovery to deal with incomplete labels; Bucak *et al*. [4] proposed an approach based on group LASSO to handle incomplete label assignments in the training data.

Multiple approaches proposed CNN-based techniques for multi-label classification [29, 34, 11]. Wu *et al*. [34] proposed a triplet loss function to draw images with similar label sets. Izadinia *et al*. [11] proposed a robust logistic loss function to train CNNs from user-provided tags. Wang *et al*. [29] proposed a recurrent neural network that predicts multiple labels one-by-one in a specified order. Other CNN-based methods handle multiple labels by treating an image as multiple images sampled from different regions, after which the problem is transformed into multiple single-label classification problems. Wei *et al*. [31] proposed a multi-object annotation framework based on object proposals, while Ren *et al*. [22] proposed a multiple instance learning framework to optimize hidden labels to region mapping.

Pairwise ranking has been applied to multi-label image classification. Weston *et al.* [32] proposed the WARP loss, and Gong *et al.* [9] applied the WARP loss to train CNNs for multi-label image annotation. Wang *et al.* [30] fused the pointwise and pairwise labeling to jointly improve the annotation and attribute prediction accuracy.

## 3. Approach

Let $\mathcal{D} = \{(x_i, Y_i)\}_{i=1}^{N}$ be our dataset with $x_i \in \mathbb{R}^d$ the $i$-th image and $Y_i \subseteq \mathcal{Y}$ the corresponding label set, where $\mathcal{Y} \triangleq \{1, 2, \cdots, K\}$ is the set of all possible labels. Each image can have different number of labels $k_i = |Y_i|$.

Our goal is to learn a multi-label image classifier $F(x)$ that not only computes the confidence scores for all labels but also determines which labels to include in the result. We decompose this into $F(x) = g(f(x))$ where $f(x) : \mathbb{R}^d \to \mathbb{R}^K$ is a label prediction model that maps an image to the $K$-dimensional label space $\mathcal{Y}$ representing the confidence scores, and $g(f(x)) : \mathbb{R}^K \to \mathbb{R}^k$ is a label decision model that produces a label set $\hat{Y} \subseteq \mathcal{Y}$ of size $k \leq K$ from the confidence scores. To take advantage of end-to-end image representation learning, we adapt a Convolutional Neural Network to our problem and focus on designing a novel loss function for $f(x)$ and an inference scheme for $g(f(x))$.

### 3.1. Label Prediction

We learn a label prediction model $f(x;\theta) \in \mathbb{R}^K$ with parameters $\theta$ by solving an optimization problem

$$\min_{\theta} \frac{1}{N} \sum_{i=1}^{N} l(f(x_i;\theta), Y_i) + \mathcal{R}(\theta) \quad (4)$$

where $l(f(x_i;\theta), Y_i)$ is a loss function and $\mathcal{R}(\theta)$ is a regularization term.

**Loss function.** One desirable property of the function $f(x)$ is that it should produce a vector whose values for true labels $Y$ are greater than those for the negative labels $\mathcal{Y} - Y$,

$$f_u(x) > f_v(x), \quad \forall u \in Y, v \notin Y \quad (5)$$

where $f_u(x)$ is the $u$-th element of $f(x)$. We can design a loss function to enforce such property within the framework of *learning to rank* [16] via pairwise comparisons,

$$l_{\text{rank}} = \sum_{v \notin Y_i} \sum_{u \in Y_i} \max\left(0, \alpha + f_v(x_i) - f_u(x_i)\right), \quad (6)$$

where $\alpha$ is a hyper-parameter that determines the margin, commonly set to 1.0 [9]. Unfortunately, the hinge function form above is non-smooth and thus is difficult to optimize.

We propose a smooth approximation to Eqn. (6) using the log-sum-exp pairwise (LSEP) function,

$$l_{\text{lsep}} = log\left(1 + \sum_{v \notin Y_i} \sum_{u \in Y_i} exp\left(f_v(x_i) - f_u(x_i)\right)\right). \quad (7)$$

Our LSEP form is, asymptotically, an upper bound of the following hinge loss form,

$$l_{\text{lsep}}^{\text{asym}} = \sum_{v \notin Y_i} \sum_{u \in Y_i} max\left(0, \alpha_i + f_v(x_i) - f_u(x_i)\right), \quad (8)$$

which, unlike Eqn. (6), allows the model to have adaptive margins per sample pair. This provides the flexibility to the learning problem. Most importantly, it is differentiable and smooth everywhere, which makes it easier to optimize. As we will show in Section 4, our LSEP form has favorable theoretical guarantees compared to the hinge alternative.

**Negative sampling.** Pairwise comparison has the $O(K^2)$ time complexity, which can cause scalability issues when the vocabulary size is large.

To make our loss function scale linearly to the vocabulary size, we adapt the negative sampling technique used in word2vec [18]. Specifically, we sample at most $t$ pairs from the Cartesian product ($t$ is empirically set as 1000). Denoting this by $\phi(Y_i; t) \subseteq Y_i \otimes (\mathcal{Y} - Y_i)$, our new loss is

$$l_{\text{lsep}} = log\left(1 + \sum_{\phi(Y_i;t)} exp\left(f_v(x_i) - f_u(x_i)\right)\right). \quad (9)$$

**Gradients.** The gradient of Eqn. (9) with respect to $f(x_i)$ can be computed as

$$\frac{\partial l_{\text{lsep}}}{\partial f(x_i)} = -\frac{1}{l_{\text{lsep}}} \sum_{\phi(Y_i;t)} \Delta Y_{i,u,v} e^{(-f(x_i) \Delta Y_{i,u,v})} \quad (10)$$

where $\Delta Y_{i,u,v} = Y_{i,u} - Y_{i,v}$ and $Y_{i,u}$ is a one-hot vector that sets all but $u$-th element of $Y_i$ to zero.

### 3.2. Comparison to Related Loss Functions

We are, of course, not the first to propose to use the pairwise ranking for multi-label classification. We compare our LSEP loss with two published techniques based on pairwise ranking: WARP loss [32, 9] and BP-MLL loss [37].

**WARP.** Extending the pairwise ranking loss in Eqn. (6), Weston *et al.* [32] proposed the WARP loss function that puts different weights on violations:

$$l_{\text{warp}} = \sum_{v \notin Y_i} \sum_{u \in Y_i} w(r_i^u) \max\left(0, \alpha + f_v(x_i) - f_u(x_i)\right), \quad (11)$$

where each pairwise violation is weighted by a monotonically increasing function $w(\cdot)$, and $r_i^u$ is the predicted rank of the positive label $u$. The intuition is that if the positive label is ranked lower, the violation should be penalized higher. While the original work by Weston *et al.* was evaluated on single-label classification, Gong *et al.* [9] successfully applied this to the multi-label image annotation task.

There are two key differences between our LSEP loss and the WARP loss. (1) Our LSEP loss is smooth everywhere,

which makes it easier to optimize. (2) It does not have the weighting function $w(r_i^u)$ of Eqn. (11).

Note that we could add the weight term in our LSEP loss. While the weight function $w(r_i^u)$ is well motivated, however, our preliminary experiments suggest that it does not provide performance boost to our model compared to not having it. We believe this is because our LSEP loss has an implicit weighting effect to penalize the lower ranked positives harder. As shown in Eqn. (10), the inner sum of the gradients of the LSEP loss is proportional to the $\Delta Y_{i,u,v}$, which means if a positive label $u$ is ranked lower ($\Delta Y_{i,u,v}$ is larger), its gradients will become larger (penalized more).

**BP-MLL.** Zhang *et al.* [37] proposed BP-MLL (back-propagation for multi-label learning), which is a multi-layer perceptron (MLP) trained with an exponential pairwise ranking loss for genomics and text categorization tasks. Using the notation in this paper, its loss has the following form,

$$l_{\text{BP-MLL}} = \sum_{v \notin Y_i} \sum_{u \in Y_i} exp(f_v(x_i) - f_u(x_i)). \quad (12)$$

which is different from our LSEP loss by missing the bias term inside the logarithmic function. This may be seen as a trivial difference, but the bias term is crucial for getting numeric stability during optimization [2]. Another difference, the iteration over the Cartesian product in Eqn. (12) makes its complexity quadratic to the vocabulary size, unlike in our case which is linear thanks to the negative sampling.

Besides the fact that our LSEP loss is easier to optimize due to the numeric stability and the negative sampling, we found that it is also optimizing a better underlying objective. To see this, the BP-MLL loss function is asymptotically equivalent to the following loss,

$$l_{\text{BP-MLL}}^{\text{asym}} = \sum_{v \notin Y_i} \sum_{u \in Y_i} (f_v(x_i) - f_u(x_i)). \quad (13)$$

Compared with the asymptotic form of our LSEP loss (Eqn. (8)), Eqn. (13) is different by how it handles the non-violating cases, i.e., $f_u(x_i) > f_v(x_i)$. Eqn. (8) maximizes the difference only up to a margin $\alpha_i$, but (13) pushes the difference to infinity. As shown in Vapnik [28], hinge loss formulations usually generalize better than least square formulations; we expect, asymptotically, the same to happen between Eqn. (8) (LSEP) and Eqn. (13) (BP-MLL). In our experiments, we empirically show that the objective represented by Eqn. (8) performs better in practice, because it makes the optimization focus on violating cases $f_v(x_i) > f_u(x_i)$.

### 3.3. Label Decision

Most existing approaches use simple heuristics for label decision, such as top-$k$ (*i.e.*, choose top $k$ results from a ranked list) or global thresholding (*i.e.*, choose labels whose confidence score is greater than a single threshold $\theta$); both methods ignore image content when making the decision.

We make our label decision model $g(\cdot)$ as a learnable function that finds the optimal decision criterion by considering the image content. In particular, we propose two versions of $g(\cdot)$, one that estimates the label count (improving top-$k$), another that estimates the optimal threshold values for each class (improving global thresholding).

We define $g(\cdot)$ as an MLP on top of $f'(x)$, which is the second to the last layer of the CNN (e.g., `fc7` layer) as input to the MLP and adding two hidden layers on top, where each hidden layer is followed by a ReLU nonlinearlity [19]. The output layer has different forms for the two versions.

**Label count estimation.** We cast the problem as $n$-way classification that estimates the number of visual concepts appearing in an image, where $n$ is the maximum number of labels permitted in the model. Any image with more than $n$ labels is capped at $n$ labels.

We define the output layer as an $n$-way softmax function, and use it to determine how many labels to return, *i.e.*, $\hat{k} = \arg \max g(f'(x))$. The top $\hat{k}$ ranked labels in $f(x)$ are included in the final output. We train the model using the softmax (multinomial logistic) loss, using $k_i = |Y_i|$ as the ground-truth number of labels for the $i$-th image:

$$l_{\text{count}} = -\log\left(\frac{\exp(g_{k_i}(f'(x_i)))}{\sum_{j=1}^{n} \exp(g_j(f'(x_i)))}\right) \quad (14)$$

where $g_j(\cdot)$ is the $j$-th element of a vector $g(\cdot)$.

**Threshold estimation.** We cast the problem as $K$-dimensional regression that estimates optimal threshold values for each class given an image. The output of the MLP is a vector of label confidence thresholds $\boldsymbol{\theta} \in \mathbb{R}^K$ for making the decision:

$$\hat{Y} = \{l | f_k(x) > \theta_k, \ \forall k \in [1, K]\}, \quad (15)$$

that is, we include labels whose confidence score is greater than the estimated threshold value. The true objective is to make an exact subset of the output, *i.e.* Eqn. (1), but we relax it via cross entropy loss:

$$l_{thresh} = -\sum_{k=1}^{K} Y_{i,k} log(\delta_\theta^k) + (1 - Y_{i,k}) log(1 - \delta_\theta^k), \quad (16)$$

where $Y_{i,k} = \{0, 1\}$ is the $k$-th label for the $i$-th sample, and $\delta_\theta$ is the sigmoid function, $sigmoid(f_l(x_i) - \theta_l)$

**Training.** To share the same image representation with the label prediction model, we fix the weights of the CNN and optimize only for the weights in the label decision model. We note that, while one could try jointly learning $f(x)$ and $g(f(x))$ by formulating an objective with a multi-task loss that combines Eqn. (9) with either Eqn. (14) or Eqn. (16), we empirically found that our sequential training approach almost always provides better performance.

### 3.4. Implementation details

For the label prediction model, we use VGG16 [26] pre-trained on the ImageNet ILSVRC challenge dataset [23] as our CNN model, replace the softmax loss from the original model with our LSEP loss, and finetune it for 10 epochs. For the label decision model, we set the maximum number of labels $n = 4$ based on our data analysis: 88.6% and 83.7% of images from NUS-WIDE and MS-COCO datasets have labels less than or equal to 4. We use 100 units in the first and 10 units in the second hidden layer. We train the MLP from scratch for 50 epochs. The regularization term in Eqn. (4) is defined as an $l_2$ norm with a weight decay of 5e-5. We optimize both models using the SGD with momentum of 0.9 and the learning rate of 0.001.

## 4. Theoretical Analysis

In this section, we show the benefit of our LSEP loss from the point of view of the Bayes consistency and risk minimization. The Bayes consistency is an important property for a loss function to achieve the right objective [5, 24].

Let's consider a Bayes prediction rule

$$f_k(x) = P(u \in Y|x), \quad (17)$$

which determines the rank of the $u$-th label in $\mathcal{Y}$. Note that $P(u \in Y|x)$ is a marginal probability over all possible subsets of $\mathcal{Y}$ that contain the $u$-th label,

$$P(u \in Y|x) = \sum_{Y \subset \mathcal{Y}: u \in Y} P(Y|x) \quad (18)$$

Below we show that the solution that minimizes our LSEP loss (Eqn. (7)) satisfies the Bayes prediction rule.

**Theorem 1.** *If $f^*(x)$ is the minimizer of Eqn.(7), then*

$$f_u^*(x) = \log P(u \in Y|x) + c, \quad \forall u \in \mathcal{Y} \quad (19)$$

*Proof.* Consider $f(x)$ that minimizes the risk,

$$R(f) = \mathbb{E}[l_{\text{lsep}}(f(x), Y)] = \int l_{\text{lsep}}(f(x), Y) \quad (20)$$

The LSEP loss is analytically equivalent with the following loss without the logarithmic,

$$l_{\exp} = \sum_{u \in Y} \sum_{v \notin Y} exp\left(-\frac{1}{2}f(x)^T \Delta Y_{u,v}\right), \quad (21)$$

where $\Delta Y_{u,v} = Y_u - Y_v$ and $Y_u$ is a one hot vector with only the $u$-th term set to one. Substituting this with $l_{\text{lsep}}$ and denoting $\gamma_{u,v} = exp\left(-\frac{1}{2}f(x)^T \Delta Y_{u,v}\right)$, we can rewrite Eqn. (20) in terms of a sample $x$ as

$$\begin{aligned} R(f|x) &= \mathbb{E}[l_{\exp}(f(x), Y) \mid x] \\ &= \sum_{Y \subset \mathcal{Y}} P(Y|x) l_{\exp}(f(x), Y) \\ &= \sum_{Y \subset \mathcal{Y}} P(Y|x) \sum_{u \in Y, v \notin Y} \gamma_{u,v} \\ &= \sum_{u,v} \sum_{Y \subset \mathcal{Y}: u \in Y, v \notin Y} P(Y|x) \gamma_{u,v} \\ &= \sum_{u,v} P(u \in Y, v \notin Y) \gamma_{u,v} \end{aligned} \quad (22)$$

The first and second derivatives of Eqn. (22) are

$$\frac{\partial R(f|x)}{\partial f(x)} = -\frac{1}{2} \sum_{u,v} \beta_{u,v} \Delta Y_{u,v}^T \gamma_{u,v} \quad (23)$$

$$\frac{\partial^2 R(f|x)}{\partial f(x)^2} = \frac{1}{4} \sum_{u,v} \beta_{u,v} \Delta Y_{u,v} \Delta Y_{u,v}^T \gamma_{u,v} \quad (24)$$

where $\beta_{u,v} = P(u \in Y, v \notin Y)$. Since the Hessian form in Eqn (24) is positive semidefinite, the risk in Eqn. (22) is convex and there is a global minimum, w.r.t. $f(x)$. Setting the derivative to zero, we find the global minimum:

$$f^*(x)^T \Delta Y_{u,v} = \log \frac{P(u \in Y, v \notin Y|x)}{P(u \notin Y, v \in Y|x)}, \forall u, v \in \mathcal{Y} \quad (25)$$

Therefore, for the optimal function $f^*(x)$, it is guaranteed that $f_u^*(x) \geq f_v^*(x)$, if and only if

$$P(u \in Y, v \notin Y|x) \geq P(u \notin Y, v \in Y|x). \quad (26)$$

Because $P(u \in Y|x, v \notin Y) = P(u \in Y|x) - P(u \in Y, v \in Y|x)$, Eqn. (26) follows $P(u \in Y|x) \geq P(v \in Y|x)$, therefore $f_u^*(x) \geq f_v^*(x)$, if and only if $P(u \in Y|x) \geq P(v \in Y|x)$, which means $f^*(x)$ implements the Bayes prediction rule in Eqn. (17). □

## 5. Experiments

We evaluate our method on the VOC2007 [8], NUS-WIDE [7] and the MS-COCO [15] datasets, comparing with several baseline methods. We also compare our LSEP loss (9) against different pairwise ranking loss functions, and discuss our design decisions for the label decision module [1].

### 5.1. Methodology

**Datasets.** We use the VOC2007 [8], NUS-WIDE [7], and the MS-COCO [15] datasets for our experiments.

The VOC2007 contains 10K images labeled with 20 common objects, and is divided into half for training and testing splits. The NUS-WIDE contains 260K images labeled with 81 visual concepts. We follow the experimental protocol in Gong *et al*. [9] and use 150K randomly sampled images for training and the rest for testing. The MS-COCO contains 120K images labeled with 80 common objects. Labels are provided for training and validation splits; we use the validation split for testing, as is common in the literature.

For NUS-WIDE and MS-COCO datasets, we discarded images whose URLs are invalid and do not have any label. As a result, for the NUS-WIDE we used 150K and 50,261 images for training and testing, respectively; for the MS-COCO we used 82,081 and 40,137 images for training and testing, respectively. On each dataset, we cross-validated model hyperparameters based on random 5% samples held out from the training set.

---
[1]The code to reproduce the results can be found here: https://bitbucket.org/raingo-ur/mll-tf

| Method | NUS-WIDE | | | | | | MS-COCO | | | | | | VOC2007 | |
|---|---|---|---|---|---|---|---|---|---|---|---|---|---|---|
| | PC-P | PC-R | OV-P | OV-R | $F_1$ | 0-1 | PC-P | PC-R | OV-P | OV-R | $F_1$ | 0-1 | $F_1$ | 0-1 |
| Softmax (K) | 42.7 | 52.5 | 54.2 | 67.5 | 43.2 | 5.02 | 56.2 | 56.8 | 59.7 | 61.7 | 54.8 | 5.63 | 73.2 | 56.6 |
| Ranking (K) | 42.6 | 56.3 | 54.7 | 68.2 | 45.1 | 5.31 | 57.0 | 57.8 | 60.2 | 62.2 | 55.4 | 5.71 | 70.8 | 56.3 |
| BP-MLL (K) | 40.9 | 56.8 | 53.9 | 67.1 | 44.0 | 4.89 | 55.8 | 56.0 | 58.9 | 60.8 | 53.6 | 5.22 | 65.3 | 54.0 |
| WARP (K) | 43.8 | <u>57.1</u> | 54.5 | 67.9 | 45.5 | 5.13 | 55.5 | 57.4 | 59.6 | 61.5 | 54.8 | 5.48 | 71.9 | <u>56.9</u> |
| Softmax ($\theta$) | 50.6 | **57.8** | 62.2 | **76.0** | 52.1 | <u>26.1</u> | 58.4 | **59.0** | 59.5 | <u>63.6</u> | 57.2 | 16.6 | 74.1 | 53.4 |
| Ranking ($\theta$) | <u>51.3</u> | 56.5 | <u>64.6</u> | <u>70.8</u> | <u>52.5</u> | 25.6 | 60.7 | 57.9 | 64.0 | 62.6 | <u>58.0</u> | <u>17.3</u> | <u>75.2</u> | 52.0 |
| BP-MLL ($\theta$) | 36.7 | 48.2 | 49.4 | 57.0 | 39.2 | 17.5 | 50.1 | 56.6 | 52.7 | 61.6 | 51.6 | 14.5 | 68.1 | 42.5 |
| WARP ($\theta$) | 48.4 | 53.1 | 59.8 | 64.6 | 48.5 | 21.3 | 57.3 | <u>58.9</u> | 60.7 | 63.5 | 56.9 | 15.9 | 74.7 | 47.5 |
| Wang et al. [29] | 40.5 | 30.4 | 49.9 | 61.7 | - | - | <u>66.0</u> | 55.6 | <u>69.2</u> | **66.4** | - | - | - | - |
| LSEP (ours) | **66.7** | 45.9 | **76.8** | 65.7 | **52.9** | **33.5** | **73.5** | 56.4 | **76.3** | 61.8 | **62.9** | **30.6** | **79.1** | **64.6** |

Table 1: **Experimental results** (K: top-$K$, $\theta$: thresholding). Our approach (using threshold estimation) achieves the state-of-the-art results on all datasets in terms of the two precision-based metrics, the $F_1$ score, and the exact match score (0-1).

**Metrics.** Similar to Gong et al. [9], we report our results in terms of per-class precision/recall (PC-P/R) and overall precision/recall (OV-P/R). The per-class measures are:

$$\text{PC-P} = \frac{1}{K}\sum_{y=1}^{K}\frac{N_y^c}{N_y^p}, \quad \text{PC-R} = \frac{1}{K}\sum_{y=1}^{K}\frac{N_y^c}{N_y^g} \quad (27)$$

where $K$ is the vocabulary size, $N_y^c$ is the number of correctly predicted images for the $y$-th label, $N_y^p$ is the number of predicted images for the $y$-th label, and $N_y^g$ is the number of ground-truth images for the $y$-th label.

Note that the per-class measures treat all classes equal regardless of their sample size, so one can obtain a high performance by focusing on getting rare classes right. To compensate this, we also measure overall precision/recall:

$$\text{OV-P} = \frac{\sum_{y=1}^{K} N_y^c}{\sum_{y=1}^{K} N_y^p}, \quad \text{OV-R} = \frac{\sum_{y=1}^{K} N_y^c}{\sum_{y=1}^{K} N_y^g} \quad (28)$$

which treats all samples equal regardless of their classes. In addition, we measure the macro $F_1$ score [36], which is an $F_1$ score averaged across all classes, and the 0/1 exact match accuracy (see Eqn. (1)), which considers a prediction correct only if all the labels are correctly predicted.

We note that our metrics are different from the standard metric for the VOC2007 dataset (mAP). This is because our approach focuses on the practical setting of multi-label image classification where the expected output of the system is a set of labels, rather than a ranked list of labels with confidence scores.

### 5.2. Baselines

Since our approach consists of two parts, label prediction (Section 3.1) and label decision (Section 3.3), we compare each part against different baseline approaches. We also compare against the state-of-the-art results from the CNN-RNN approach proposed by Wang et al. [29].

**Label prediction.** We compare our LSEP loss against four approaches: softmax, the standard pairwise ranking (6), WARP (11), and BP-MLL (12).

Although the softmax functions are originally proposed for single-label classification, in practice it is often used for the multi-label scenario via problem transformation [21]. Specifically the softmax loss can be adapted to the multi-label scenario as,

$$l_{\text{softmax}} = \sum_{y \in Y_i} \log\left(\frac{exp(f_y(x_i))}{\sum_{j \in \mathcal{Y}} exp(f_j(x_i))}\right). \quad (29)$$

**Label decision.** We evaluate two common label decision approaches as baselines: top-$k$ and thresholding. Instead of choosing an arbitrary number (e.g., $k = 3$, $\theta = 0.5$), we cross-validated the optimal values, choosing the parameter that achieved the best result on the validation split in terms of the $F_1$ score; we chose the $F_1$ score because it is a balanced measure between precision and recall. For top-$k$, we varied $k = 1 : 10$; and for thresholding, we varied $\theta$ among 50 equal-interval values between the minimum and the maximum values of predicted confidence scores.

### 5.3. Results and discussion

**Overall system performance.** Table 1 shows our evaluation results on the VOC2007 [8], NUS-WIDE [7] and the MS-COCO [15] datasets. We report our method with a label decision module that estimates the optimal per-class threshold values.

To the best of our knowledge, our result is the best reported performance in the literature in terms of the two precision-based metrics (PC-P, OV-P), and the $F_1$ score. Notably, our approach outperforms all baselines by a large margin in terms of the exact match score (0-1). It is the most strict measure of all for multi-label classification and highlights the superiority of our method in the practical setting.

| Method | NUS-WIDE | | | | | | MS-COCO | | | | | | VOC2007 | |
|---|---|---|---|---|---|---|---|---|---|---|---|---|---|---|
| | PC-P | PC-R | OV-P | OV-R | $F_1$ | 0-1 | PC-P | PC-R | OV-P | OV-R | $F_1$ | 0-1 | $F_1$ | 0-1 |
| Top-$k$ | 44.8 | 55.6 | 54.8 | 68.3 | 45.5 | 5.39 | 56.2 | 58.6 | 60.5 | 62.4 | 55.8 | 6.19 | 72.5 | 57.6 |
| Threshold | 55.0 | **57.0** | 67.2 | **73.4** | **55.0** | 29.3 | 59.0 | **63.4** | 61.5 | **67.1** | 59.8 | 23.2 | 76.4 | 54.5 |
| Label count est. | 61.4 | 46.1 | 73.7 | 64.7 | 50.9 | 33.4 | 67.7 | 57.6 | 72.0 | 62.2 | 61.4 | 30.4 | 78.3 | **66.3** |
| Thresh. est. (ours) | **66.7** | 45.9 | **76.8** | 65.7 | 52.9 | **33.5** | **73.5** | 56.4 | **76.3** | 61.8 | **62.9** | **30.6** | **79.1** | 64.6 |

Table 2: Comparison of different label decision approaches. We use our LSEP loss for label prediction. The second two rows show label count estimation and per-class threshold estimation (ours).

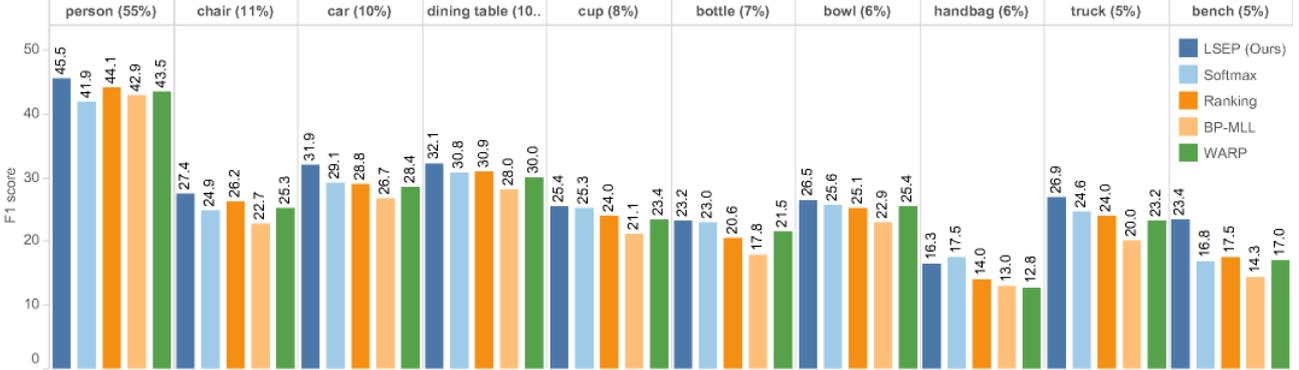

Figure 2: Per-class $F_1$ scores on top 10 most frequent classes in the MS-COCO dataset. Our method is based on the per-class threshold estimation, baseline methods are based on thresholding.

Figure 2 shows per-class $F_1$ scores on the top 10 frequent classes in the MS-COCO dataset; again, our method is based on the optimal per-class threshold estimation, and baseline methods are based on thresholding. It shows our method performing consistently better than baselines across classes.

Overall, we notice that our approach performs particularly well in terms of precision; this allow us to achieve the best $F_1$ scores even with the low recall rates. Regarding label decision baselines, we notice that thresholding outperforms top-$k$ inference for the most cases. This is because thresholding is less restrictive than top-$k$; the latter is forced to always include $k$ labels in the output, even if there is a single visual concept appearing in an image.

**Label prediction comparison.** Figure 3 shows average precision-recall (PR) curves on the two datasets. The PR curves allow us to tease apart the effect of label prediction without considering label decision. The figure shows that our LSEP loss outperforms baselines across various ranges of decision points, suggesting the robustness of our LSEP loss compared to the baselines.

**Label decision comparison.** Table 2 compares different label decision approaches using our LSEP loss based label prediction as the base model. Similar to our observation from Table 1, the threshold-based method outperforms its top-$k$ counterpart in terms of precision and the exact match score. This suggests that the current common practice of choosing top-$k$ labels in multi-label classification can be improved simply by using a thresholding mechanism.

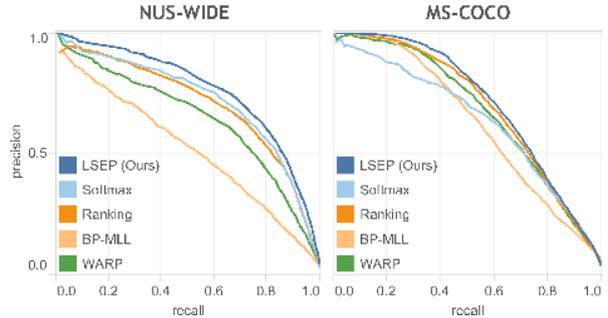

Figure 3: Average precision-recall curves on both datasets.

Also, our per-class threshold estimation outperforms the simple thresholding mechanism by a large margin in terms of precision and the exact match score. This shows the benefit of *learning* the optimal threshold values for each class category, instead of using one global threshold value.

**Qualitative analysis.** Our label decision model estimates per-class thresholds given an image. We showed that this greatly improves the performance. One byproduct of the model is that it naturally estimates the complexity of recognizing visual concepts in a given image. Figure 4 shows images from each class ordered by their estimated confidence thresholds. We can see that the task difficulty increases as the estimated thresholds increases. This effect is caused by how we train the label decision model: To reduce mistakes made by the label prediction model, the label decision model is forced to be more conservative (*i.e.*, need a high threshold value) for less obvious images, and vice versa. This is reflected in the images shown in Figure 4, *e.g.*, the two rightmost images of 'Dog' depicts a dog riding a motorbike and surfing on the water; these cases are quite rare and thus more likely to be mistaken.

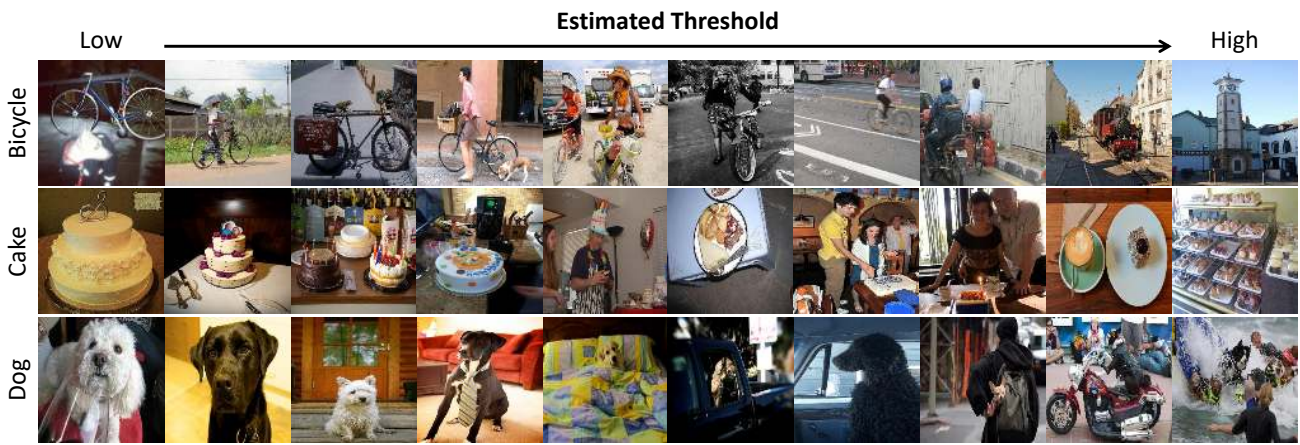

Figure 4: Estimated confidence threshold values correlate with the task difficulty. Shown here are randomly sampled images from each class ordered by their estimated threshold values. As the threshold value increases, it becomes more difficult to recognize the corresponding visual concept in a given image. All images are from the MS-COCO dataset [15].

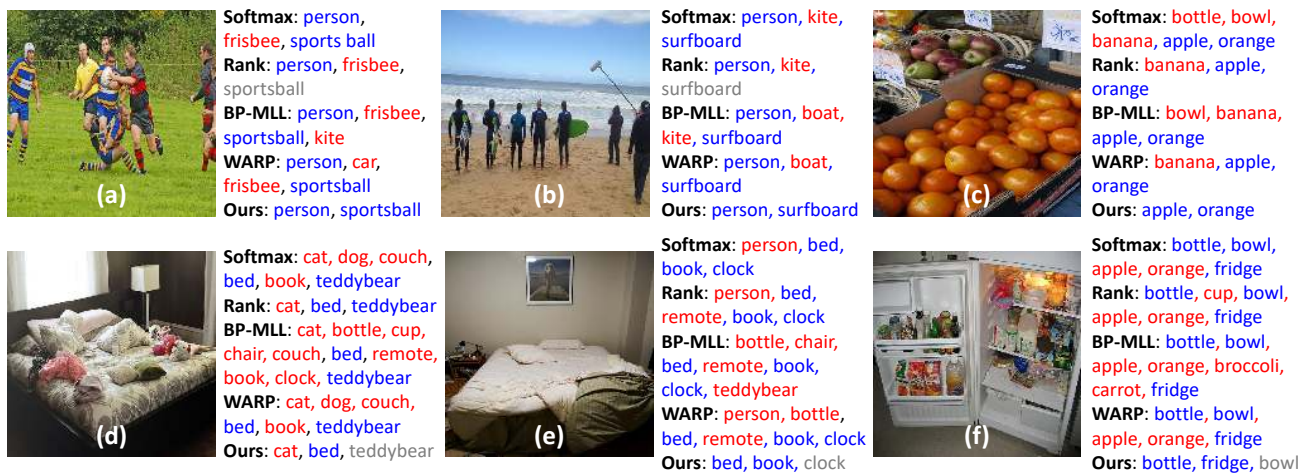

Figure 5: **Qualitative results** (top: success cases, bottom: failure cases, blue: true positive, red: false positive, gray: false negative). Our approach tends to be more conservative in label decision, including only highly relevant labels in the output.

Figure 5 shows qualitative results. As shown in Table 1, our approach tends to be more conservative in label decision, producing labels that contain only relevant visual concepts (this behavior is in line with the high precision of our approach). For example, Figure 5 (a-c) shows our approach producing exact match labels, while Figure 5 (d-f) shows our approach missing some labels that are hard to find in the images, rather than including incorrect ones.

## 6. Conclusion

We proposed two techniques to improve existing pairwise ranking based multi-label image classification: a novel pairwise ranking loss function that is easier to optimize; and a label decision model that determines which labels to include in the output. We discussed the superiority of our LSEP loss function via theoretical analysis, and demonstrated our approach on the VOC2007, NUS-WIDE and the MS-COCO datasets, achieving the best reported results in the literature in terms of precision and $F_1$ score.

Our work focused on improving existing ranking-based approaches to multi-label classification. In the future, we would like to explore ways to leverage the distinctive properties of multi-label problem, such as label dependency, label sparsity, and missing labels.

## References

[1] A. Alessandro, G. Corani, D. Mauá, and S. Gabaglio. An ensemble of bayesian networks for multilabel classification. In *IJCAI*, 2013. 1, 2

[2] S. Bach, A. Binder, G. Montavon, F. Klauschen, K.-R. Müller, and W. Samek. On pixel-wise explanations for non-linear classifier decisions by layer-wise relevance propagation. *PloS one*, 10(7), 2015. 4

[3] M. R. Boutell, J. Luo, X. Shen, and C. M. Brown. Learning multi-label scene classification. *Pattern recognition*, 37(9):1757–1771, 2004. 1

[4] S. S. Bucak, P. K. Mallapragada, R. Jin, and A. K. Jain. Efficient multi-label ranking for multi-class learning: application to object recognition. In *CVPR*, 2009. 2

[5] W. Cheng, E. Hüllermeier, and K. J. Dembczynski. Bayes optimal multilabel classification via probabilistic classifier chains. In *ICML*, 2010. 5

[6] T.-S. Chua, H.-I. Pung, G.-J. Lu, and H.-S. Jong. A concept-based image retrieval system. In *HICSS*, 1994. 1

[7] T.-S. Chua, J. Tang, R. Hong, H. Li, Z. Luo, and Y.-T. Zheng. Nus-wide: A real-world web image database from national university of singapore. In *CIVR*, 2009. 1, 2, 5, 6

[8] M. Everingham, L. Van Gool, C. K. Williams, J. Winn, and A. Zisserman. The pascal visual object classes (voc) challenge. *IJCV*, 2010. 2, 5, 6

[9] Y. Gong, Y. Jia, T. Leung, A. Toshev, and S. Ioffe. Deep convolutional ranking for multilabel image annotation. *arXiv preprint arXiv:1312.4894*, 2013. 1, 3, 5, 6

[10] D. Hsu, S. Kakade, J. Langford, and T. Zhang. Multi-label prediction via compressed sensing. In *NIPS*, 2009. 1, 2

[11] H. Izadinia, B. C. Russell, A. Farhadi, M. D. Hoffman, and A. Hertzmann. Deep classifiers from image tags in the wild. In *MMCommons*, 2015. 2

[12] A. Kapoor, R. Viswanathan, and P. Jain. Multilabel classification using bayesian compressed sensing. In *NIPS*, 2012. 1, 2

[13] A. Krizhevsky, I. Sutskever, and G. E. Hinton. Imagenet classification with deep convolutional neural networks. In *NIPS*, 2012. 1

[14] A. Lambrecht and C. Tucker. When does retargeting work? information specificity in online advertising. *Journal of Marketing Research*, 50(5), 2013. 1

[15] T.-Y. Lin, M. Maire, S. Belongie, J. Hays, P. Perona, D. Ramanan, P. Dollár, and C. L. Zitnick. Microsoft coco: Common objects in context. In *European Conference on Computer Vision*, pages 740–755. Springer, 2014. 2, 5, 6, 8

[16] T.-Y. Liu. Learning to rank for information retrieval. *FTIR*, 3(3), 2009. 3

[17] W. Liu and I. Tsang. Large margin metric learning for multi-label prediction. In *AAAI*, 2015. 2

[18] T. Mikolov, I. Sutskever, K. Chen, G. S. Corrado, and J. Dean. Distributed representations of words and phrases and their compositionality. In *NIPS 2013*. 3

[19] V. Nair and G. E. Hinton. Rectified linear units improve restricted boltzmann machines. In *ICML*, 2010. 4

[20] J. Nam, J. Kim, E. Loza Mencía, I. Gurevych, and J. Fürnkranz. Large-scale multi-label text classification — revisiting neural networks. In *ECML PKDD 2014*. 2

[21] J. Read. Scalable multi-label classification. 2010. 1, 2, 6

[22] Z. Ren, H. Jin, Z. Lin, C. Fang, and A. Yuille. Multi-instance visual-semantic embedding. *arXiv preprint arXiv:1512.06963*, 2015. 2

[23] O. Russakovsky, J. Deng, H. Su, J. Krause, S. Satheesh, S. Ma, Z. Huang, A. Karpathy, A. Khosla, M. Bernstein, A. C. Berg, and L. Fei-Fei. ImageNet Large Scale Visual Recognition Challenge. *IJCV*, 2015. 1, 5

[24] M. J. Saberian and N. Vasconcelos. Multiclass boosting: Theory and algorithms. In *NIPS*, 2011. 5

[25] X. Shu, D. Lai, H. Xu, and L. Tao. Learning shared subspace for multi-label dimensionality reduction via dependence maximization. *Neurocomputing*, 2015. 1, 2

[26] K. Simonyan and A. Zisserman. Very deep convolutional networks for large-scale image recognition. In *ICLR*, 2015. 5

[27] Y. Song, D. McDuff, D. Vasisht, and A. Kapoor. Exploiting sparsity and co-occurrence structure for action unit recognition. In *FG*, 2015. 1, 2

[28] V. N. Vapnik and V. Vapnik. *Statistical learning theory*, volume 1. Wiley New York, 1998. 4

[29] J. Wang, Y. Yang, J. Mao, Z. Huang, C. Huang, and W. Xu. CNN-RNN: A unified framework for multi-label image classification. In *CVPR*, 2016. 1, 2, 6

[30] Y. Wang, S. Wang, J. Tang, H. Liu, and B. Li. Ppp: Joint pointwise and pairwise image label prediction. In *Proceedings of the IEEE Conference on Computer Vision and Pattern Recognition*, pages 6005–6013, 2016. 3

[31] Y. Wei, W. Xia, J. Huang, B. Ni, J. Dong, Y. Zhao, and S. Yan. CNN: single-label to multi-label. *CoRR*, abs/1406.5726, 2014. 2

[32] J. Weston, S. Bengio, and N. Usunier. Wsabie: Scaling up to large vocabulary image annotation. In *IJCAI*, 2011. 1, 3

[33] B. Wu, S. Lyu, and B. Ghanem. "ml-mg: Multi-label learning with missing labels using a mixed graph". In *ICCV 2015*. 1, 2

[34] F. Wu, Z. Wang, Z. Zhang, Y. Yang, J. Luo, W. Zhu, and Y. Zhuang. Weakly semi-supervised deep learning for multi-label image annotation. *IEEE Transactions on Big Data*, 1(3):109–122, 2015. 2

[35] X. Yang, Y. Li, and J. Luo. Pinterest board recommendation for twitter users. In *ACM Multimedia*, 2015. 1

[36] Y. Yang. An evaluation of statistical approaches to text categorization. *Information retrieval*. 6

[37] M.-L. Zhang and Z.-H. Zhou. Multilabel neural networks with applications to functional genomics and text categorization. *TKDE*, 2006. 3, 4

[38] M.-L. Zhang and Z.-H. Zhou. Ml-knn: A lazy learning approach to multi-label learning. *Pattern recognition*, 2007. 2

[39] F. Zhao and Y. Guo. Semi-supervised multi-label learning with incomplete labels. In *AAAI*, 2015. 1, 2